\documentclass[letterpaper, 10 pt, conference]{ieeeconf}  

\IEEEoverridecommandlockouts                              

\usepackage{graphics} 
\usepackage{epsfig} 
\usepackage{mathptmx} 
\usepackage{times} 
\usepackage{amsmath} 
\usepackage{amssymb}  

\usepackage[dvipsnames, table]{xcolor}
\usepackage{glossaries}
\usepackage{hyperref}
\usepackage{booktabs}
\usepackage{fontawesome5}
\usepackage{cuted}
\usepackage{caption}
\usepackage{subcaption}
\usepackage{multirow}
\usepackage[noabbrev, capitalise]{cleveref}
\usepackage{tikz}
\usetikzlibrary{spy}
\usepackage{pgfplots}
\usepgfplotslibrary{colormaps}
\pgfplotsset{compat=1.18}

\usepackage{subcaption}
\usepackage{array}
\usepackage{xspace} 
\usepackage{tabularx}
\usepackage[per-mode=symbol, group-digits=integer, detect-all=true, input-symbols={()}]{siunitx}
\usetikzlibrary{positioning, arrows.meta}
\usepackage{amsmath,bm}
\usepackage{subcaption}
\usepackage[normalem]{ulem}
\newcommand{\MethodName}{DFI-OmniStereo}

\newcommand{\cf}{\textit{cf.}\xspace}
\newcommand{\etal}{\textit{et al.}\xspace}

\newcolumntype{Y}{>{\centering\arraybackslash}X} 

\definecolor{ouryellow}{RGB}{255,176,0}
\definecolor{ourorange}{RGB}{254,97,0}
\definecolor{ourpink}{RGB}{220,38,127}
\definecolor{ourpurple}{RGB}{120,94,240}
\definecolor{ourblue}{RGB}{100,143,255}

\tikzstyle{process_module} = [rectangle, rounded corners, minimum width=3cm, minimum height=1.5cm, text centered, draw=black, text=white, fill=orange, line width=0.5mm]
\tikzstyle{process_image} = [minimum width=3cm, minimum height = 1.5 cm, inner sep=0, outer sep=0]
\tikzstyle{arrow} = [line width=2mm, ->, >=stealth, draw=orange]

\newacronym{vit}{ViT}{Vision Transformer~\cite{dosovitskiy2020image}}
\newacronym{dpt}{DPT}{Dense Prediction Transformer}
\newacronym{cnn}{CNN}{Convolutional Neural Network}
\newacronym{cgru}{ConvGRU}{Convolutional Gated Recurrent Unit}
\newacronym{cgev}{CGEV}{Combined Geometry Encoding Volume}
\newacronym{igev}{IGEV-Stereo}{Iterative Geometry Encoding Volume for Stereo Matching}

\usepackage{etoolbox}
\makeatletter
\patchcmd{\@makecaption}
  {\scshape}
  {}
  {}
  {}
\patchcmd{\@makecaption}
  {\\}
  {:\ }
  {}
  {}
\makeatother

\title{\LARGE \bf
Boosting Omnidirectional Stereo Matching\\with a Pre-trained Depth Foundation Model
}
\usepackage{fancyhdr}
\usepackage{setspace}
\fancypagestyle{firststyle}
{

	\fancyhf{}
	\lfoot{{\footnotesize\begin{spacing}{.5}\parbox{\linewidth}{\vspace{-1.85em}%
					\textcolor{gray}{To appear in \emph{Proceedings of the IEEE/RSJ International Conference on Intelligent Robots and Systems}, Hangzhou, China, 2025.%
					\\\hrule\vspace{\baselineskip}
					\copyright~2025 IEEE. Personal use of this material is permitted. Permission from IEEE must be obtained for all other uses, in any current or future media, including reprinting/republishing this material for advertising or promotional purposes, creating new collective works, for resale or redistribution to servers or lists, or reuse of any copyrighted component of this work in other works.} 
	}\end{spacing}}}
}

\author{
Jannik Endres\textsuperscript{1,2}\qquad\quad
Oliver Hahn\textsuperscript{2}\qquad\quad
Charles Corbi\`ere\textsuperscript{1}\\
Simone Schaub-Meyer\textsuperscript{2,3}\qquad\quad
Stefan Roth\textsuperscript{2,3}\qquad\quad
Alexandre Alahi\textsuperscript{1} \\[4pt]
\scriptsize{\url{https://vita-epfl.github.io/DFI-OmniStereo-website}}
\thanks{\textsuperscript{1}\'Ecole Polytechnique F\'ed\'erale de Lausanne (EPFL) \quad
\textsuperscript{2}Technical University of Darmstadt, Department of Computer Science \quad
\textsuperscript{3}hessian.AI}
}
\begin{document}
\maketitle
\thispagestyle{firststyle}
\pagestyle{plain}
\bstctlcite{BSTcontrol}


\begin{abstract}
Omnidirectional depth perception is essential for mobile robotics applications that require scene understanding across a full 360° field of view.
Camera-based setups offer a cost-effective option by using stereo depth estimation to generate dense, high-resolution depth maps without relying on expensive active sensing.
However, existing omnidirectional stereo matching approaches achieve only limited depth accuracy across diverse environments, depth ranges, and lighting conditions, due to the scarcity of real-world data.
We present {\MethodName}, a novel omnidirectional stereo matching method that leverages a large-scale pre-trained foundation model for relative monocular depth estimation within an iterative optimization-based stereo matching architecture.
We introduce a dedicated two-stage training strategy to utilize the relative monocular depth features for our omnidirectional stereo matching before scale-invariant fine-tuning.
{\MethodName} achieves state-of-the-art results on the real-world Helvipad dataset, reducing disparity MAE by approximately 16\% compared to the previous best omnidirectional stereo method.
\end{abstract}
\section{INTRODUCTION}

\begin{figure}
    \centering
    \includegraphics[width=\columnwidth]{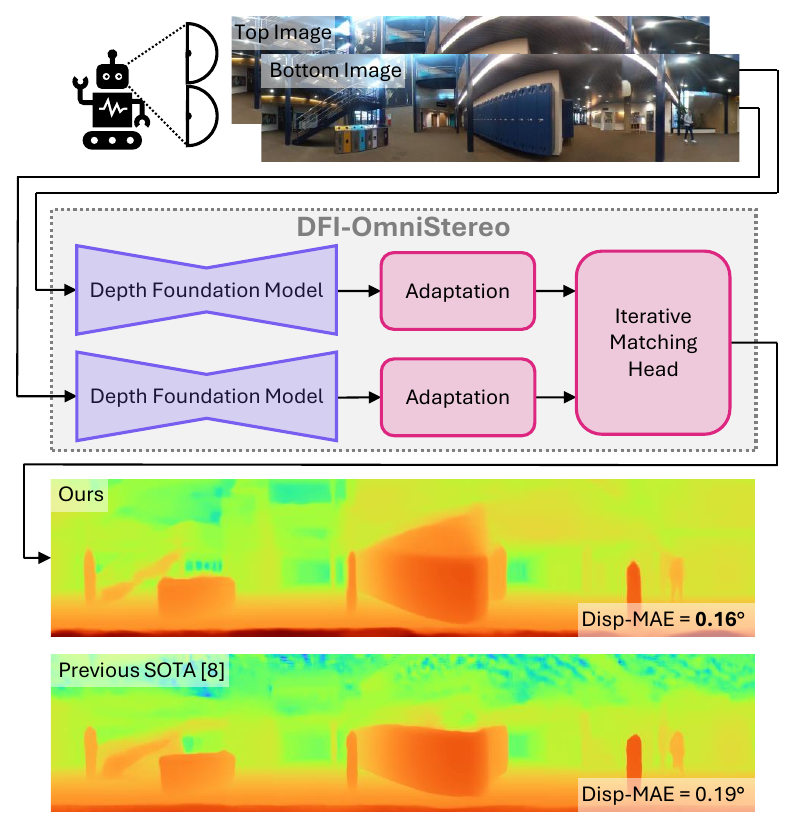}    
    \caption{
    \textbf{Overview of our proposed omnidirectional stereo matching approach {\MethodName}}.
    Given a pair of equirectangular images captured by two vertically stacked omnidirectional cameras, our method integrates a large-scale pre-trained monocular relative depth foundation model into an iterative stereo matching approach. {\MethodName} improves disparity and depth estimation accuracy, significantly outperforming the previous state-of-the-art method.
    We visualize predicted disparity on a log-scale (red indicates high disparity and low depth; vice versa for blue).
    \label{fig:summary}}
    \vspace{-1em}
\end{figure}

Mobile robots are increasingly being deployed across various domains, including agriculture~\cite{oliveira2021advances}, autonomous driving~\cite{yurtsever2020survey}, healthcare~\cite{holland2021service}, search and rescue missions~\cite{tong2024robots}, and warehouse automation~\cite{sodiya2024ai}. In these applications, accurate depth perception is crucial to construct reliable 3D representations of a robot’s environment to achieve essential tasks such as path planning, mapping, and manipulation. Traditionally, LiDAR sensors have been the preferred choice for acquiring depth information due to their high precision and 360° field of view. However, they are often prohibitively expensive and provide relatively sparse measurements. These drawbacks have motivated the exploration of more cost-effective approaches, such as camera-based configurations.

Omnidirectional stereo depth estimation \cite{won2019omnimvs,jiang2024romnistereo,zayene2024helvipad,Chen2023UnsupervisedOE,won2020end} has recently raised significant interest as it overcomes the narrow field of view of conventional stereo matching. While recent work \cite{zayene2024helvipad} has begun to mitigate the scarcity of real-world data via a novel dataset of 360° image pairs, current omnidirectional methods still face challenges in generalizing across diverse environments. Meanwhile, monocular depth estimation has seen remarkable progress, driven by the introduction of depth foundation models such as Depth Anything~\cite{yang2024depth}, which are trained on vast amounts of both labeled and unlabeled data. Our work aims to leverage the strengths of these large-scale pre-trained models to enhance stereo matching in omnidirectional systems.

In this paper, we introduce {\MethodName} (\textit{\underline{D}epth \underline{F}oundation Model-based \underline{I}terative \underline{Omni}directional \underline{Stereo} Matching}), a novel omnidirectional stereo matching method that integrates a pre-trained monocular depth foundation model into an iterative optimization-based stereo matching architecture, as illustrated in \cref{fig:summary}. Our approach follows a two-stage training strategy in which \emph{(i)} the stereo matching head learns to adapt to the new feature space and camera setup while keeping the foundation model fixed, and then \emph{(ii)} the foundation model’s decoder is unfrozen to be fine-tuned using a scale-invariant loss without foregoing its generalization capabilities.
We conduct extensive experiments to compare {\MethodName} with other omnidirectional stereo matching methods, perform detailed analyses, and evaluate its training sample efficiency as well as its generalization capabilities.

Specifically, our contributions are as follows:
\textbf{(1)} We leverage a large-scale pre-trained monocular depth foundation model as a feature extractor integrated into an iterative optimization-based stereo matching architecture.
\textbf{(2)} We design a two-stage training strategy to adapt monocular foundation model features to the omnidirectional stereo matching setup.
We partially fine-tune the foundation model, and employ a scale-invariant error in log space (SILog loss) for stereo matching.
\textbf{(3)} Our method demonstrates state-of-the-art results on the Helvipad~\cite{zayene2024helvipad} dataset, a challenging real-world benchmark for omnidirectional stereo matching. In addition, {\MethodName} shows generalization capabilities to other datasets and high training sample efficiency.
\section{RELATED WORK}
\paragraph{Stereo matching}
A disparity map can be estimated from two images and then deterministically converted to depth assuming a calibrated stereo setup.
Early deep learning methods \cite{vzbontar2016stereo} use \glspl{cnn} for matching cost calculation or end-to-end disparity estimation \cite{mayer2016large}.
More recent 3D networks \cite{kendall2017end, chang2018pyramid, zhang2019ga, guo2019group} introduce 4D cost volumes via feature concatenation and employ a 3D encoder-decoder design for context aggregation.
Many recent works in deep stereo matching follow two paradigms: \gls{vit}-based methods and iterative optimization-based methods \cite{tosi2024survey}.
GMStereo~\cite{xu2023unifying} and CroCo-Stereo~\cite{weinzaepfel2023croco} propose a unified transformer-based architecture for optical flow and stereo.
RAFT-Stereo~\cite{lipson2021raft} presents an iterative multi-resolution approach that refines disparity maps using a multi-level 3D correlation volume and a learned context encoding via convolutional GRUs~\cite{cho2014propertiesneuralmachinetranslation}.
Following RAFT-Stereo, many subsequent architectures \cite{li2022practical, zhao2023high, wang2024selective, chen2024mocha} propose an iterative refinement of the disparity.
In particular, IGEV-Stereo~\cite{xu2023iterative} constructs a 3D cost volume that fuses local cues from a multi-level 3D correlation volume with global context from a regularized 4D cost volume to iteratively refine disparity estimates.
Vankadar \etal~\cite{vankadari2024dusk} and ViTAStereo~\cite{li2024roadformer} use features from the vision foundation models DINO~\cite{caron2021emerging} and DINOv2~\cite{oquab2024dinov2} for stereo matching.
Our approach extends this family by incorporating a large-scale pre-trained depth foundation model as feature extractor to consider the strong correlation between relative depth and disparity.\footnote{
Note that, independently and concurrently with our work, Wen~\etal~\cite{wen2025stereo} and Cheng~\etal~\cite{cheng2025monster} adopt a conceptually similar methodology to \MethodName{}, applying it to conventional stereo matching.
We also fine-tune the decoder of a foundation model to (omnidirectional) stereo matching.
}

\paragraph{Omnidirectional depth estimation}
Three-dimensional scene geometry is inferred from data captured over a 360° field of view in omnidirectional depth estimation.
Several works \cite{tateno2018distortion, won2019sweepnet, pintore2021slicenet} approach this task by applying an equirectangular projection to map the spherical field of view of omnidirectional cameras onto a plane. However, this results in distortions at the top and bottom of the image \cite{zelnik2005squaring}.
To address these distortions, some works adapt \glspl{cnn} to spherical images \cite{su2017learning, cohen2018spherical, tateno2018distortion}, while others develop distortion-aware \glspl{vit} \cite{shen2022panoformer, yun2023egformer, carlsson2024heal}.
A second stream of works employs alternative projection methods \cite{cheng2018cube, wang2020bifuse, eder2020tangent}, such as cubemaps \cite{cheng2018cube}.
Most omnidirectional stereo matching methods \cite{won2019omnimvs, jiang2024romnistereo, Chen2023UnsupervisedOE, won2020end, won2019sweepnet, komatsu2020360, meuleman2021real} rely on four fisheye cameras capturing images from different directions with overlapping fields of view, enabling stereo matching.
However, all these methods validate their results exclusively on synthetic datasets.
A simpler and more cost-effective setup uses just two omnidirectional cameras, a top and a bottom one. However, until recently, research has been limited by the lack of large-scale datasets \cite{zayene2024helvipad}.
To the best of our knowledge, the only existing architectures for omnidirectional stereo matching with this configuration  are 360SD-Net~\cite{wang2020360sd} and 360-IGEV-Stereo~\cite{zayene2024helvipad}.
360SD-Net, built on PSMNet~\cite{chang2018pyramid}, addresses distortions by encoding a polar angle image and concatenating it with the image features.
A learnable vertical shifting filter is used to adjust for varying pixel step sizes in the cost volume construction.
360-IGEV-Stereo, built on IGEV-Stereo, integrates an encoded polar angle map into its feature and context networks.
Additionally, it applies circular padding~\cite{wang2018omnidirectional} before inference to exploit the circular boundary conditions and constructs cost volumes via vertical instead of horizontal warping to suit the camera setup.
However, this previous approach yields inaccurate results near object boundaries under diverse lighting conditions (\cf \cref{fig:qualitative}) due to the limited robustness of its feature network.

\paragraph{Monocular relative depth estimation}
The goal of monocular relative depth estimation is to predict scale- and shift-invariant depth from a single RGB image.
Early work on depth estimation focuses on metric depth, initially using hand-crafted features \cite{saxena2005learning, saxena2008make3d} and later learned deep representations \cite{eigen2014depth, liu2015deep,laina2016deeper}, but cannot generalize to multiple datasets.
MiDaS~\cite{ranftl2020towards} combines data from multiple sources by converting ground truth to scale- and shift-invariant values.
Consequently, this model is capable of cross-dataset generalization.
In later MiDaS versions \cite{birkl2023midas}, the \gls{dpt} decoder~\cite{ranftl2021vision} converts the token-based feature representation of the \gls{vit}~\cite{dosovitskiy2020image} encoder into image-like feature representations at multiple resolutions, which are combined into a final dense prediction.
Recent approaches develop foundational models for monocular depth estimation by further scaling the architecture and training data.
Depth Anything~\cite{yang2024depth} leverages self-supervised DINOv2~\cite{oquab2024dinov2} image feature representations and the \gls{dpt} decoder.
Depth Anything is trained with a teacher-student approach using unlabeled images with pseudo-labels from the teacher.
Depth Anything V2~\cite{yang2024depth2} extends its predecessor by replacing the labeled real-world data with synthetic images and increasing the size of the teacher model.
In this paper, we show that the large-scale pre-training performed to create foundational models for monocular relative depth estimation can be effectively leveraged when addressing specialized challenges such as omnidirectional stereo matching.

\begin{figure*}[ht]
\centering
\includegraphics[width=\textwidth]{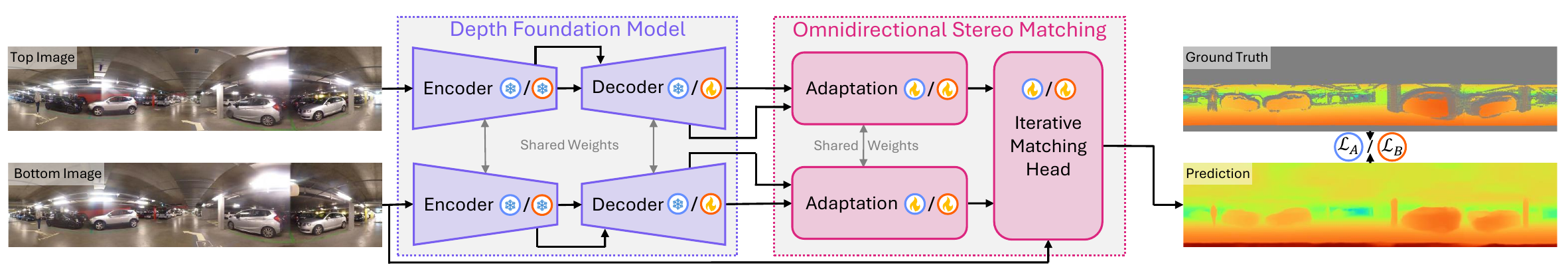}
\caption{
\textbf{Overview of {\MethodName}}.
A shared depth foundation model (\textcolor{ourpurple}{purple}) is utilized to extract representations from a top and bottom image.
Subsequently, an omnidirectional stereo matching head (\textcolor{ourpink}{pink}) predicts disparity, utilizing the image features as follows:
The intermediate representations and relative depth maps of both images are adapted to be processed as multi-scale feature maps by the iterative matching head.
This head predicts a disparity map using vertical warping for cost volume construction.
The training consists of two stages.
In training stage A (\textcolor{ourblue}{blue}), we adapt the stereo matching head to the omnidirectional data and the foundation model features (foundation model frozen) using a conventional stereo matching loss $\mathcal{L}_{A}$.
In stage B (\textcolor{ourorange}{orange}), we fine-tune the foundation model decoder and the stereo matching head, utilizing a scale-invariant logarithmic loss \( \mathcal{L}_{B} \).
Frozen and trainable modules are denoted with a snowflake and fire symbol, respectively.
\label{fig:architecture_detail}}
\vspace{-1.5em}
\end{figure*}

\section{\MethodName}
We present {\MethodName}, an end-to-end model for omnidirectional stereo matching.
Given an omnidirectional stereo image pair $(I_\text{t}, I_\text{b})$, consisting of a top and bottom image, each of dimensions $\mathbb{R}^{H \times W \times 3}$, our goal is to predict the vertical disparity between the two images.
We demonstrate how a foundation model for relative monocular depth estimation can be leveraged and adapted for the targeted task of omnidirectional stereo matching (\cref{subsec:architecture}). 
We propose a two-stage training strategy by first aligning the two building blocks of our method, the foundation model and the stereo matching head. Next, we can leverage the large-scale pre-trained foundation model and fine-tune it in conjunction with the stereo matching head (\cref{subsec:training}).
\cref{fig:architecture_detail} provides an overview of our proposed method.

\subsection{{\MethodName} Architecture \label{subsec:architecture}}
Our proposed framework consists of two core parts, a depth foundation model to extract features and an omnidirectional stereo matching head.

\paragraph{Depth foundation model as feature extractor}
For feature extraction, we propose to leverage a recently published foundation model for relative depth estimation, specifically Depth Anything V2~\cite{yang2024depth2}.
This foundation model consists of a \gls{vit} encoder~\cite{dosovitskiy2020image} and a \acrfull{dpt} decoder~\cite{ranftl2021vision} to predict dense relative depth from the ViT patch encodings.
Depth Anything V2 has been trained on millions of synthetic and pseudo-labeled real images.
Due to the strong task relationship between relative depth and disparity estimation, these learned features provide a good starting point for stereo matching.
Specifically, we use the predicted relative depth map as well as feature maps at the four intermediate resolutions of the decoder as input for omnidirectional stereo matching.

\paragraph{Omnidirectional stereo matching}
The omnidirectional stereo matching head predicts the disparity values based on the information provided by the feature extractor.
Our omnidirectional stereo matching head is inspired by IGEV-Stereo~\cite{xu2023iterative}, an established, iterative optimization-based stereo matching architecture.
However, as in our case the feature extractor, i.e., depth foundation model, and the stereo matching head have been developed and trained initially for different application scenarios, the intermediate features of the foundation model are not usable out of the box as multi-scale feature inputs for stereo matching.
To bridge this gap, our adaptation module employs bilinear interpolation to align the spatial dimensions, as well as a learnable linear layer to adjust the number of channels.
Originally, IGEV-Stereo encodes the input images with an additional, small encoder network and concatenates these features
with the extracted features of the feature extractor at a resolution of $(H/4,W/4)$. 
Instead, we encode the relative depth map of each image to leverage the similarity between relative depth and disparity.
To finally predict the disparity from these features, we mostly follow previous work~\cite{xu2023iterative} by using 3D and 4D group-wise correlation volumes \cite{guo2019group}. 
Given our omnidirectional stereo setup with a top and bottom image, we shift the top image vertically to construct the correlation volumes as opposed to horizontally in the typical case of a left-right image pair.
Analogously to other stereo matching methods \cite{lipson2021raft, xu2023iterative}, we iteratively update an initial disparity estimate at $(H/4, W/4)$ resolution based on the volumes and multi-scale context features, extracted from the bottom image by a \gls{cnn}, using multilevel convolutional GRUs.
For upsampling the disparity to the full image resolution, we follow previous work \cite{xu2023iterative} of guided upsampling.
Different from \cite{xu2023iterative}, we incorporate the encoding of the relative depth map instead of the bottom image encoding alongside the context network features in this process, leveraging the similarity between relative depth and disparity boundaries.

\subsection{Training Strategy \label{subsec:training}}
Due to the limited availability of omnidirectional stereo images with ground-truth depth labels, we leverage pre-trained modules for our framework, although trained originally on different data (monocular and stereo rectilinear images).
To effectively combine the two modules as well as realize the benefit of our omnidirectional setting without losing the generalization capabilities of the foundation model, we propose a two-stage training strategy besides the architectural adaptations.
In each stage, we employ a loss that penalizes all intermediate disparity predictions with exponentially increasing weights, as it is common in iterative stereo matching methods \cite{lipson2021raft, li2022practical, zhao2023high}.

\paragraph{Stage A -- Feature adaptation}
The goal of the first training stage is the adaptation of the stereo matching head to the new image feature representation, the camera setup, and the omnidirectional imagery.
Consequently, we train the stereo matching and adaptation components of {\MethodName} while keeping the foundation model frozen.
In this stage, we apply a widely used stereo matching loss function.
Analogous to \cite{xu2023iterative}, we incorporate a smooth L1 loss term \( \mathcal{L}_{sL_1} \),
for the initial disparity estimate to enhance robustness against outliers.
With $N$ disparity updates, the L1-based loss for the first training stage $\mathcal{L}_{A}$ is defined as 
\begin{equation}
    \mathcal{L}_{A}(\{\boldsymbol{\hat{d}}_i\}^N_{i=0}) = \mathcal{L}_{sL_1}\left(\boldsymbol{\hat{d}}_0, \boldsymbol{d}\right) + \sum_{i = 1}^{N} \gamma^{N - i} \mathcal{L}_{L_1}\left(\boldsymbol{\hat{d}}_i, \boldsymbol{d}\right),
    \label{eq:LA}
\end{equation}
where \( \boldsymbol{\hat{d}}_i \) denotes the predicted disparities at iteration \( i \) for pixels with ground truth, and \( \boldsymbol{d} \) represents the corresponding ground-truth disparity values.
$\gamma$ is an attenuation factor.
For a ground-truth disparity map with $n$ valid pixels, the L1 loss $\mathcal{L}_{L_1}$ and smooth L1 loss $\mathcal{L}_{sL_1}$ are defined as 
\begin{equation}
    \mathcal{L}_{L_1}\left(\boldsymbol{\hat{d}}, \boldsymbol{d}\right) = \frac{\| \boldsymbol{\hat{d}} - \boldsymbol{d} \|_1}{n},
\end{equation}
\begin{equation}
    \mathcal{L}_{sL_1}\left(\boldsymbol{\hat{d}}, \boldsymbol{d}\right) = 
    \frac{1}{n} \sum_{j = 1}^{n} \ell_{sL_1}\left(\hat{d}_{j}, d_{j}\right),
\end{equation}
where
\begin{equation}
    \ell_{sL_1}\left(\hat{d}, d\right) =
    \begin{cases} 
        \frac{\left(\hat{d} - d\right)^2}{2}, & \text{if } \left|\hat{d} - d\right| < 1, \\
        \scriptstyle \left|\hat{d} - d\right| - \frac{1}{2}, & \text{otherwise}.
    \end{cases}
\end{equation}

\paragraph{Stage B -- Scale-invariant fine-tuning}
Subsequently, we want to further fine-tune the foundation model to the omnidirectional imagery and the task of stereo matching.
To retain the foundation model's high quality feature representations obtained in the extensive pre-training, we solely fine-tune the decoder.
We train the stereo matching head in this training stage as well.
Note that the learning rate applied to the foundation model decoder is significantly lower than the learning rate of the stereo matching head.
We utilize the scale-invariant error in log space (SILog loss) \( \mathcal{L}_{SIL} \), as introduced by \cite{eigen2014depth}.
This loss does not penalize incorrect estimates of the log-depth up to an unknown scale factor and weights small and large depth values more equally by taking the logarithm.
Given the employed dataset's diverse depth ranges, spanning both indoor and outdoor scenes, we convert the SILog loss \( \mathcal{L}_{SIL} \) into an iterative variant \( \mathcal{L}_{B} \) for the second training stage, analogous to \cref{eq:LA}, to prevent the model from overfitting to specific depth scales:
\begin{equation}
    \mathcal{L}_{B}\left(\{\boldsymbol{\hat{d}}_i\}^N_{i=0}\right) = \mathcal{L}_{SIL}\left(\boldsymbol{\hat{d}}_0, \boldsymbol{d}\right) + \sum_{i = 1}^{N} \gamma^{N - i} \mathcal{L}_{SIL}\left(\boldsymbol{\hat{d}}_i, \boldsymbol{d}\right),
\end{equation}
where
\begin{equation}
    \mathcal{L}_{SIL}\bigl(\boldsymbol{\hat{d}}, \boldsymbol{d}\bigr) = \frac{1}{n} \sum_{j = 1}^{n}\!\delta_{\log}\!\left(\hat{d}_j, d_{j} \right)^2-\frac{\lambda}{n^2}\!\left(\sum_{j = 1}^{n} \!\delta_{\log}\!\left(\hat{d}_j, d_{j}\right)\!\right)^{\!\!2}
\end{equation}
with \( \delta_{log}\left(\hat{d},d\right) = \log \hat{d} - \log d \) and $\lambda$ being a tuning parameter \cite{eigen2014depth}.
To the best of our knowledge, we are the first to leverage a SILog loss specific to iterative stereo matching.

\section{EXPERIMENTS}
In this section, we compare {\MethodName} against existing stereo-matching methods on real-world data and provide insights into its accuracy.
We further study the model's training sample-efficiency, in particular in low training data regimes, and its generalization capabilities.

\subsection{Dataset and Evaluation Metrics}
\label{subsec:dataset_metrics}

\paragraph{Dataset} We train and evaluate our approach on Helvipad~\cite{zayene2024helvipad}, the only real-world omnidirectional stereo depth estimation dataset with a top-bottom camera setup.
While there exist synthetic datasets, Stereo-MP3D~\cite{wang2020360sd,Matterport3D} and Stereo-SF3D~\cite{wang2020360sd,armeni2017joint}, which share our configuration, their lack of photorealism makes them unsuitable for training and evaluation.
Wang \etal~\cite{wang2020360sd} additionally provide three real-world images without ground truth, which we include in a qualitative analysis in \cref{subsec:zero_shot}.
The Helvipad dataset consists of 27K training, 3K validation, and 10K test image pairs, along with ground-truth depth and disparity maps.
Each split features a mix of indoor scenes, outdoor daytime, and outdoor nighttime scenes.
Following \cite{zayene2024helvipad}, the task is defined as estimating disparity \( d=\theta_b-\theta_t \) from a top-bottom stereo image pair, where \( \theta_t \) and \( \theta_b \) are the polar angles of the top and bottom cameras' spherical camera model.
According to \cite{zayene2024helvipad} the disparity $d$ can be converted to depth \( r_b \) using
\begin{equation}
     r_b = B \left(\frac{\sin(\theta_b)}{\tan(d)} + \cos(\theta_b)\right),
\end{equation}
with \( B \) denoting the baseline between the cameras.

\begin{table*}[t]
\vspace{0.35em}
    \centering
    \caption{\textbf{Comparative results of omnidirectional stereo depth estimation on the Helvipad~\cite{zayene2024helvipad} test split.} Comparing \MethodName\;to existing stereo matching approaches on both disparity and depth metrics (MAE, RMSE, MARE, and LRCE). Lower values (\textdownarrow) indicate better results.}
    \setlength{\tabcolsep}{2pt}
    \sisetup{table-number-alignment=center}
    \newcommand{\mytablecolumnwidth}{0.07\textwidth}
    \renewcommand{\arraystretch}{0.87}
    \footnotesize
        \begin{tabularx}{\textwidth}{
        >{\hspace{-\tabcolsep}\raggedright\columncolor{white}[\tabcolsep][\tabcolsep]}
        lY
        S[table-format=1.3, table-column-width=\mytablecolumnwidth]
        S[table-format=1.3, table-column-width=\mytablecolumnwidth]
        S[table-format=1.3, table-column-width=\mytablecolumnwidth]
        S[table-format=1.3, table-column-width=\mytablecolumnwidth]
        S[table-format=1.3, table-column-width=\mytablecolumnwidth]
        S[table-format=1.3, table-column-width=\mytablecolumnwidth]
        S[table-format=1.3, table-column-width=\mytablecolumnwidth]
        S[table-format=1.3, table-column-width=\mytablecolumnwidth]}
            \toprule
            & & \multicolumn{4}{c}{\textbf{Disparity} (°)} & \multicolumn{4}{c}{\textbf{Depth} (m)} \\
            \cmidrule(lr){3-6} \cmidrule(lr){7-10}
            \multirow{-2}{*}{\vspace{0.5em}\textbf{Method}} & \multirow{-2}{*}{\vspace{0.5em}\textbf{Stereo Setting}} & \text{MAE $\downarrow$} & \text{RMSE $\downarrow$} & \text{MARE $\downarrow$} & \text{LRCE $\downarrow$} & \text{MAE $\downarrow$} & \text{RMSE $\downarrow$} & \text{MARE $\downarrow$} & \text{LRCE $\downarrow$}  \\
            \midrule
            PSMNet~\cite{chang2018pyramid} & Conventional & 0.286 & 0.496 & 0.248 & {--} & 2.509 & 5.673 & 0.176 & 1.809 \\
            360SD-Net~\cite{wang2020360sd} & Omnidirectional & 0.224 & 0.419 & 0.191 & {--} & 2.122 & 5.077 & 0.152 & 0.904 \\
            IGEV-Stereo~\cite{xu2023iterative} & Conventional & 0.225 & 0.423 & 0.172 & {--} & 1.860 & 4.474 & 0.146 & 1.203 \\
            360-IGEV-Stereo~\cite{zayene2024helvipad} & Omnidirectional & 0.188 & 0.404 & 0.146 & \bfseries 0.054 & 1.720 & 4.297 & 0.130 & \bfseries 0.388 \\
            \midrule
            \rowcolor{gray!25} {\MethodName} & Omnidirectional & \bfseries 0.158 & \bfseries 0.338 & \bfseries 0.120 & 0.058 & \bfseries 1.463 & \bfseries 3.767 & \bfseries 0.108 & 0.397 \\
            \bottomrule
        \end{tabularx}
    \label{tab:comparative_results}
\end{table*}
\begin{figure*}[t]
    \centering
    \input{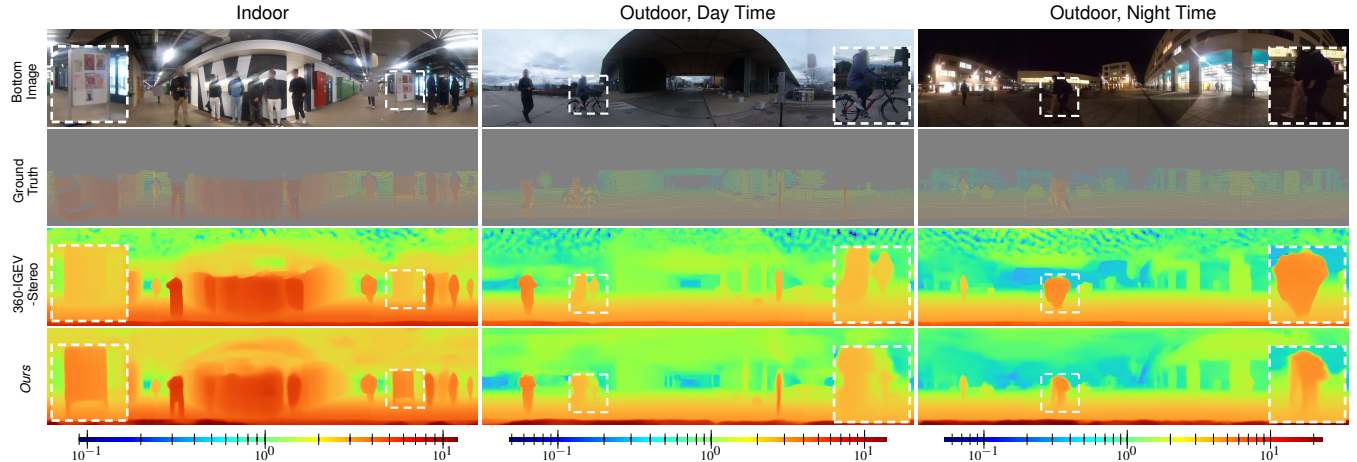}
    \vspace{-0.75em}
    \caption{\textbf{Qualitative comparison on the Helvipad~\cite{zayene2024helvipad} test split.} We visualize the bottom image, ground-truth disparity maps (°), and the predicted disparity maps (°) of the previous state-of-the-art method, 360-IGEV-Stereo, and of {\MethodName}.
    }
    \label{fig:qualitative}
\end{figure*}

\paragraph{Metrics} Following \cite{zayene2024helvipad}, we utilize the Mean Absolute Error (MAE), Root Mean Squared Error (RMSE), and Mean Absolute Relative Error (MARE) metrics for the evaluation of both disparity and depth.
Additionally, we employ the Left-Right Consistency Error (LRCE)~\cite{shen2022panoformer}, a metric tailored to omnidirectional imagery, which assesses the consistency of predictions between the left and right vertical borders of the image. Given a dataset with $M$ image pairs and the number of valid rows in the $m$-th image being $K_m$, LRCE is defined as:
\begin{equation}
\text{LRCE} = \frac{1}{M}\sum_{m=1}^{M} \frac{1}{K_m} \sum_{k=1}^{K_m} \left| \Delta \hat{y}_{m,k} - \Delta y_{m,k} \right|,
\label{eq:lrce}
\end{equation}
where $\Delta \hat{y}_{m,k} = \hat{y}_{m,k}^L - \hat{y}_{m,k}^R$ and $\Delta y_{m,k} = y_{m,k}^L - y_{m,k}^R$. Here, $\hat{y}_{m,k}$ and $y_{m,k}$ represent the predicted values and ground truth (in terms of either disparity or depth), and the superscripts $L$ and $R$ indicate the left and right image boundaries, respectively. Note, we use the depth-completed ground truth to calculate the LRCE.
A row is considered valid when a ground-truth label is provided for both the left and right sides of the image.

\subsection{Implementation Details \label{subsec:implementation_details}}
We implement our model in PyTorch~\cite{paszke2019pytorch}, building upon the codebases of Depth Anything and IGEV-Stereo.
To stabilize training, the disparity $d_\text{deg}$ in degrees is clamped at each iteration within the range $[d_\text{deg, min}, d_\text{deg, max}]$ according to the dataset statistics.
We convert angular disparity to pixel units to enable accurate warping of the top image when constructing the cost volume, setting the maximum disparity to 128 pixels.  
Training is performed at full resolution (512$\times$1920) using photometric augmentations.
For initialization, the unmodified components of the stereo matching head are initialized with IGEV-Stereo weights obtained from Scene Flow pre-training~\cite{mayer2016large}, while the modified components are initialized randomly. The image encoder \gls{vit} and the \gls{dpt} decoder are initialized using the Depth-Anything-V2-Base checkpoint.
Stage A uses a batch size of $2$ for $20$ epochs with a learning rate of $2e^{-4}$.
Subsequently, stage B uses a batch size of $1$ for $12$ epochs with a learning rate of $2e^{-5}$. Additionally, we reduce the learning rate for the foundation model weights by a factor of $50$.
We set $\gamma = 0.9$ and $\lambda = 0.15$, following \cite{yang2024depth, lipson2021raft}.
Following the recommendations in \cite{zayene2024helvipad}, we use the depth-completed ground truth for training and evaluate all metrics except for the LRCE with the original sparse ground truth. Furthermore, we apply circular padding by $64$ pixels during inference. All experiments are conducted on a single NVIDIA A100 GPU.
In the absence of computational optimizations, \MethodName{} requires \SI{0.1}{\second} per image for feature extraction, and \SI{0.2}{\second} for iterative matching, yielding a total inference time of \SI{0.4}{\second} per image.

\subsection{Comparison to State of the Art}
We compare our method, both quantitatively and qualitatively, to several state-of-the-art stereo matching methods, including PSMNet~\cite{chang2018pyramid}, 360SD-Net~\cite{wang2020360sd}, IGEV-Stereo~\cite{xu2023iterative}, and 360-IGEV-Stereo~\cite{zayene2024helvipad}.
\paragraph{Quantitative results}
\cref{tab:comparative_results} reports the comparative results on the Helvipad test split. {\MethodName} outperforms the considered methods across nearly all evaluation metrics. Notably, our method achieves the lowest disparity MAE (0.158°), which is significantly lower than the next best result, 360-IGEV-Stereo (0.188°). The left-right consistency metric LRCE of our method is marginally worse than the best-performing baseline (360-IGEV-Stereo).
Overall, the quantitative results of \MethodName{} demonstrate that foundation model features can be leveraged to reduce errors and improve disparity and depth accuracy in omnidirectional stereo matching on diverse real-world data.

\paragraph{Qualitative results} We show disparity predictions of our proposed method \MethodName\ and the previous best method 360-IGEV-Stereo in \cref{fig:qualitative}. In particular, we can observe that 360-IGEV-Stereo exhibits artifacts with low disparity values near the top of all three scenes.
Helvipad does not provide ground-truth annotations in such regions (during training), suggesting that 360-IGEV-Stereo is unable to generalize to scene structures that have not been labeled in the training data.
Overall, \MethodName\ yields predictions with sharper edges, finer details, and smoother surfaces.
For example, in the leftmost images, only {\MethodName} successfully captures the legs of the human in the foreground and the poster on the right.
Similarly, it delineates poles more accurately in the middle examples.
Under low light (right image), {\MethodName} better distinguishes the two humans in the middle of the scene.
Overall, these qualitative results demonstrate that {\MethodName} improves depth differentiation for objects, especially humans, and the background across various scenes compared to previous approaches.

\begin{table}[t]
\vspace{0.35em}
\centering
\caption{\textbf{{\MethodName} component analysis} by selectively training different architectural components during the two training stages. All remaining components are frozen.
We use the following abbreviations: FE (feature encoder), FD (feature decoder), and OS (omnidirectional stereo matching).
$^\dagger$ indicates an adjusted learning rate due to unstable training.
}
\renewcommand{\arraystretch}{0.87}
\footnotesize
\setlength\tabcolsep{2pt}
\sisetup{table-number-alignment=center}
\newcommand{\mytablecolumnwidth}{0.06\textwidth}
\begin{tabularx}{\columnwidth}{
    >{\hspace{-\tabcolsep}\raggedright\columncolor{white}[\tabcolsep][\tabcolsep]}
    cY
    S[table-format=1.3, table-column-width=\mytablecolumnwidth]
    S[table-format=1.3, table-column-width=\mytablecolumnwidth]
    S[table-format=1.3, table-column-width=\mytablecolumnwidth]
    S[table-format=1.3, table-column-width=\mytablecolumnwidth]}
\toprule
\multicolumn{2}{c}{\textbf{Trained Components}} & \multicolumn{2}{c}{\textbf{Disparity} (°)} & \multicolumn{2}{c}{\textbf{Depth} (m)} \\
\cmidrule(lr){1-2} \cmidrule(lr){3-4} \cmidrule(lr){5-6}
    Stage A & Stage B & \text{MAE~$\downarrow$} & \text{MARE~$\downarrow$} & \text{MAE~$\downarrow$} & \text{MARE~$\downarrow$} \\
\midrule
\phantom{$^\dagger$}FD+OS$^\dagger$ & {--} & 0.169 & 0.137 & 1.589 & 0.115 \\
OS & {--} & 0.165 & 0.129 & 1.508 & 0.112 \\
OS & FE+FD+OS & 0.164 & 0.131 & 1.618 & 0.115 \\
\rowcolor{gray!25} OS & FD+OS & \bfseries 0.158 & \bfseries 0.120 & \bfseries 1.463 & \bfseries 0.108 \\
\bottomrule
\end{tabularx}
\label{tab:training_strategies}
\vspace{-1em}
\end{table}

\subsection{Analyzing \MethodName \label{subsec:ablation_studies}}
\paragraph{Component training analysis} We conduct an in-depth investigation on training different components of {\MethodName} and applying different loss terms, ultimately leading to our proposed two-stage training strategy.
\cref{tab:training_strategies} summarizes experiments where we selectively train or freeze specific modules during Stage A and B.
Training the feature decoder and the omnidirectional stereo matching components together in Stage A requires a reduced learning rate to prevent loss divergence.
However, this reduction significantly deteriorates the MAE and MARE for both disparity and depth.
This suggests that the stereo matching head needs to be adapted to the new camera setup and feature space before fine-tuning the feature representation, demonstrating the importance of a two-stage training strategy.
When fine-tuning both the feature encoder and decoder, we observe a decrease in feature quality, indicated by an increase in depth MAE from $1.463 \, \mathrm{m}$ to $1.618 \, \mathrm{m}$.

\paragraph{Loss function analysis}
We analyze the impact of applying different loss terms in the two training stages.
As shown in \cref{tab:losses}, choosing the L1-based loss leads to the best results during the first training stage. Building on the better configuration, the SILog loss proves to be superior in the second training stage to adapt the relative depth representations to the omnidirectional imagery.
These findings indicate the importance of starting with an L1-based loss before transitioning to a scale-invariant loss.

\begin{table}[t]
\vspace{0.35em}
\centering
\caption{\textbf{{\MethodName} loss analysis.} We explore the impact of using different loss terms across the two training stages. Stage B 
uses the best setup (L1-based) of stage A.}
\renewcommand{\arraystretch}{0.87}
\footnotesize
\setlength\tabcolsep{2pt}
\sisetup{table-number-alignment=center}
\newcommand{\mytablecolumnwidth}{0.06\textwidth}
\begin{tabularx}{\columnwidth}{>{\hspace{-\tabcolsep}\raggedright\columncolor{white}[\tabcolsep][\tabcolsep]} lYS[table-format=1.3, table-column-width=\mytablecolumnwidth]S[table-format=1.3, table-column-width=\mytablecolumnwidth]S[table-format=1.3, table-column-width=\mytablecolumnwidth]S[table-format=1.3, table-column-width=\mytablecolumnwidth]}
\toprule
& & \multicolumn{2}{c}{\textbf{Disparity} (°)} & \multicolumn{2}{c}{\textbf{Depth} (m)} \\
\cmidrule(lr){3-4} \cmidrule(lr){5-6}
\multirow{-2}{*}{\textbf{Training Stage}\vspace{0.5em}} & \multirow{-2}{*}{\textbf{Loss}\vspace{0.5em}} & \text{MAE~$\downarrow$} & \text{MARE~$\downarrow$} & \text{MAE~$\downarrow$} & \text{MARE~$\downarrow$} \\
\midrule
&  SILog & 0.182 & 0.130 & 1.519 & 0.118 \\
\multirow{-2}{*}{Stage A} & \cellcolor{gray!25} L1-based & \cellcolor{gray!25} \bfseries 0.165 & \cellcolor{gray!25} \bfseries 0.129 & \cellcolor{gray!25} \bfseries 1.508 & \cellcolor{gray!25} \bfseries 0.112 \\
\midrule
& \cellcolor{gray!25} SILog & \cellcolor{gray!25} \bfseries 0.158 & \cellcolor{gray!25} \bfseries 0.120 & \cellcolor{gray!25} \bfseries 1.463 & \cellcolor{gray!25} \bfseries 0.108 \\
\multirow{-2}{*}{Stage B} & L1-based & 0.160 & 0.126 & 1.494 & 0.110 \\
\bottomrule
\end{tabularx}%
\label{tab:losses}
\vspace{-0.5em}
\end{table}

\paragraph{Comparison to monocular depth estimation models}
In \cref{tab:monocular_depth}, we compare \MethodName{} to the monocular depth estimation methods EGformer~\cite{yun2023egformer}, Depth Anything~\cite{yang2024depth}, and Depth Anything V2~\cite{yang2024depth2}.
EGformer~\cite{yun2023egformer} is a monocular relative depth estimation approach specialized for omnidirectional imagery.
We adapt EGformer for metric depth estimation by replacing the final activation, removing the scale-and-shift alignment of the loss, and adjusting the spherical grid to match the Helvipad images. 
Despite being designed for equirectangular images, EGformer performs significantly worse than \MethodName{} across all metrics.
Out of the box, the Depth Anything model generalizes very poorly to omnidirectional images.\footnote{\scriptsize{We use Depth Anything V1 here, since checkpoints for the ZoeDepth~\cite{bhat2023zoedepth} head fine-tuned for metric depth prediction are not available for V2. However, Yang~\etal~\cite{yang2024depth2} show that this difference should not significantly impact accuracy.}}
In addition, we extensively fine-tune Depth Anything V2 alongside the ZoeDepth~\cite{bhat2023zoedepth} metric depth prediction head on the Helvipad dataset.
\MethodName{} is consistently better across all metrics on the Helvipad dataset, highlighting the benefit of leveraging stereo cues in combination with the depth foundation model features.
We further analyze where \MethodName{} improves over Depth Anything V2 by comparing disparity MAE across three depth intervals in \cref{tab:depth_ranges}.
Each interval comprises around one third of the available ground-truth values.
Notably, our method performs particularly well at medium depth ranges (\SI{4}{m} -- \SI{9}{m}) with a disparity MAE reduction of $8.7\%$.

\begin{table}[t]
\centering
\caption{\textbf{Comparison of {\MethodName} to monocular depth estimation models.} All models are adapted to metric depth estimation.
$\ast$ refers to our adapted version for metric depth estimation.
$\dagger$ indicates the use of the large-scale pre-trained checkpoint that includes the metric depth prediction head.
$\ddagger$ refers to training using a ZoeDepth head~\cite{bhat2023zoedepth} for metric depth estimation following \cite{yang2024depth}.
}
\renewcommand{\arraystretch}{0.87}
\footnotesize
\setlength\tabcolsep{2pt}
\begin{tabularx}{\columnwidth}{X
S[table-format=1.3]
S[table-format=1.3]
S[table-format=1.3]
S[table-format=1.3]}
\toprule
& \multicolumn{2}{c}{\textbf{Disparity} (°)} & \multicolumn{2}{c}{\textbf{Depth} (m)} \\
\cmidrule(lr){2-3} \cmidrule(lr){4-5}
\multirow{-2}{*}{\textbf{Model}\vspace{0.5em}} & \text{MAE~$\downarrow$} & \text{MARE~$\downarrow$} & \text{MAE~$\downarrow$} & \text{MARE~$\downarrow$} \\
\midrule
EGformer$^\ast$ \cite{yun2023egformer} & 0.214 & 0.157 & 01.835 & 0.144 \\
Depth Anything$^\dagger$ \cite{yang2024depth} & 1.062 & 0.977 & 5.057 & 0.392 \\
Depth Anything V2$^\ddagger$ \cite{yang2024depth2} & 0.164 & 0.123 & 1.467 & 0.110 \\
\rowcolor{gray!25} {{\MethodName}} & \bfseries 0.158 & \bfseries 0.120 & \bfseries 1.463 & \bfseries 0.108 \\

\bottomrule
\end{tabularx}
\label{tab:monocular_depth}
\vspace{-1.5em}
\end{table}

\begin{table}[t]
\vspace{0.35em}
\footnotesize
\centering
\caption{\textbf{Comparing {\MethodName} to Depth Anything for different depth ranges.}
We explore the disparity MAE (in °) for three depth ranges  (in \SI{}{m}).
}
\renewcommand{\arraystretch}{0.87}
\setlength\tabcolsep{5pt}
\begin{tabularx}{\columnwidth}{X
S[table-format=1.3]
S[table-format=1.3]
S[table-format=1.3]}
\toprule
\textbf{Model} & 

\text{\SI{0}{m} - \SI{4}{m}} & \text{\SI{4}{m} - \SI{9}{m}} & \text{\SI{9}{m} - \SI{230}{m}} \\
\midrule
Depth Anything V2$^\ddagger$ \cite{yang2024depth2} & 0.182 & 0.149 & \bfseries 0.148 \\
\rowcolor{gray!25} {{\MethodName}} & \bfseries 0.181 & \bfseries 0.136 & 0.150 \\

\bottomrule
\end{tabularx}
\label{tab:depth_ranges}
\vspace{-1em}
\end{table}

\subsection{Training Sample-efficient Learning\label{subsec:sample_efficiency}}
\begin{figure}[t]
    \setlength\tabcolsep{2pt}
    \begin{tabularx}{\columnwidth}{@{}cc@{}}
    \begin{tikzpicture}
    \begin{axis}[
      width=4.9cm, height=4cm,
      title={Disparity MAE},
      ylabel={Disp-MAE ($^\circ$)},
      xmode=log,
      xtick={1,2,5,10,20,50,100},
      xticklabels={1,2,5,10,20,50,100},
      xlabel={Sampling ratio (\%)},
      ymin=0.15, ymax=0.22,
      grid=both,
      tick label style={font=\tiny},
      xlabel style={font=\tiny, yshift=2pt},
      ylabel style={font=\tiny, yshift=-4pt},
      title style={font=\footnotesize, yshift=-6pt},
      legend columns=1,
      legend cell align=left,
      legend style={
        font=\tiny,
        draw=none,
        inner sep=1pt,
        outer sep=0pt,
        fill=white,
        nodes={align=left, inner sep=0.5pt},
        at={(0.6,0.75)}, anchor=south,
        row sep=1pt,
      },
      legend image post style={xscale=0.8, yscale=0.9},
    ]
    \addplot[gray, dashed, line width=1pt] coordinates {(1,0.188) (100,0.188)};
    \addlegendentry{360-IGEV-Stereo (100\%)}
    
    \addplot[ourpurple, mark=*] coordinates {
     (1,0.213) (2,0.201) (5,0.182) (10,0.171) (20,0.163) (50,0.162) (100,0.158)};
    \addlegendentry{DFI-OmniStereo (Ours)}
    \end{axis}
    \end{tikzpicture}
    
    &    
    \begin{tikzpicture}
    \begin{axis}[
      width=4.9cm, height=4cm,
      title={Disparity MARE},
      ylabel={Disp-MARE},
      xmode=log,
      xtick={1,2,5,10,20,50,100},
      xticklabels={1,2,5,10,20,50,100},
      xlabel={Sampling ratio (\%)},
      ymin=0.12, ymax=0.17,
      grid=both,
      tick label style={font=\tiny},
      xlabel style={font=\tiny, yshift=2pt},
      ylabel style={font=\tiny, yshift=-4pt},
      title style={font=\footnotesize, yshift=-6pt},
      legend columns=1,
      legend cell align=left,
      legend style={
        font=\tiny,
        draw=none,
        inner sep=1pt,
        outer sep=0pt,
        fill=white,
        nodes={align=left, inner sep=0.5pt},
        at={(0.6,0.75)}, anchor=south,
        row sep=1pt,
      },
      legend image post style={xscale=0.8, yscale=0.9},
    ]
    \addplot[gray, dashed, line width=1pt] coordinates {(1,0.146) (100,0.146)};
    \addlegendentry{360-IGEV-Stereo (100\%)}
    
    \addplot[ourpink, mark=triangle*] coordinates {
     (1,0.163) (2,0.152) (5,0.138) (10,0.127) (20,0.122) (50,0.123) (100,0.120)};
    \addlegendentry{DFI-OmniStereo (Ours)}
    \end{axis}
    \end{tikzpicture}
    \end{tabularx}
    
    \vspace{-0.5em}
    \caption{\textbf{Training sample-efficient learning analysis} using {\MethodName} on the Helvipad dataset~\cite{zayene2024helvipad}. The training data for our method is a randomly sampled subset.
    360-IGEV-Stereo~\cite{zayene2024helvipad} is visualized as the dashed line using $100\%$ of the training data for comparison.
    }
    \label{fig:sample_efficiency}
    \vspace{-1em}
\end{figure}
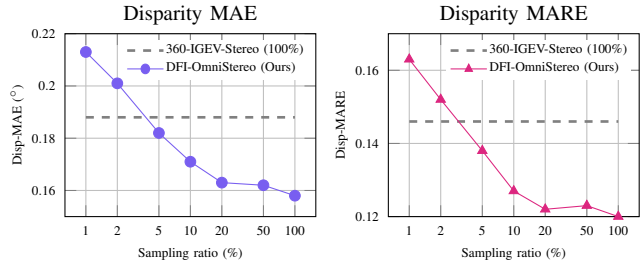

Collecting labeled real-world data is expensive. Having methods that can learn from a small amount of training samples is essential for real-world applications.
\Cref{fig:sample_efficiency} shows that \MethodName{} already achieves a lower disparity MAE than 360-IGEV-Stereo (with $100\%$ training data) when only $5\%$ of the training data, randomly sampled, are available to our model.
This highlights how leveraging large-scale pre-training from a foundation model significantly reduces the need for task-specific data.

\subsection{Qualitative Generalization Analysis\label{subsec:zero_shot}}
\begin{figure}[bt]
    \centering
    \input{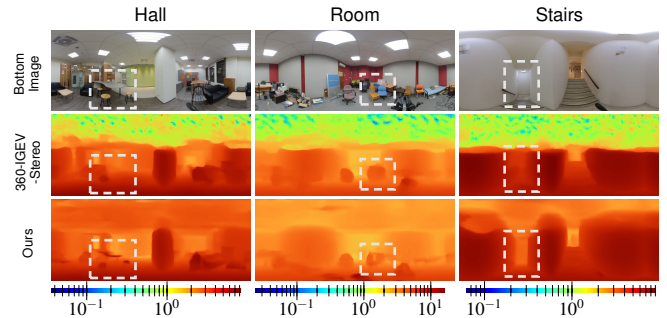}
    \vspace{-0.5em}
    \caption{
    \textbf{Qualitative comparison of generalization to real-world images from \cite{wang2020360sd}.} We visualize the bottom image and the disparity prediction (°) of 360-IGEV-Stereo and {\MethodName} (from top to bottom) using the hall, room, and stairs scene (from left to right).
    }
    \label{fig:zero_shot}
    \vspace{-1em}
\end{figure}

Finally, we aim to assess the generalization capabilities of our method.
360SD-Net~\cite{wang2020360sd} is the only work beyond Helvipad providing real-world data with a similar top-bottom camera setup.
The authors provide three scenes without ground-truth annotations.
We qualitatively analyze the cross-dataset generalization of {\MethodName} comparing to 360-IGEV-Stereo in \cref{fig:zero_shot}.
Fine details, such as the chairs and table on the left of the hall scene, are only recognizable in {\MethodName}'s disparity map.
Our method predicts the depth boundaries of objects more accurately.
For example, the brown chair and the blue desk chair in the room scene are only distinguishable in {\MethodName}'s prediction.
Homogeneous surfaces, such as the wall behind the left stairs in the stairs scene, are better visible in {\MethodName}'s disparity prediction.
These results demonstrate promising generalization capabilities of our method {\MethodName} when transferring to new cameras and stereo baselines.

\section{LIMITATIONS AND FUTURE WORK}
While \MethodName{} achieves state-of-the-art metric results, we did not focus on efficiency and real-time capabilities. 
Future work could address this through model compression techniques, knowledge distillation into a smaller task-specific foundation model component, or by replacing the iterative stereo matching.
We rely on the only real-world omnidirectional stereo depth dataset with a top-bottom camera setup, so we test the large pre-trained depth model in a single specialized setting. Future work should evaluate its generalization to other data-scarce stereo scenarios, such as aerial, or underwater imaging.

\section{CONCLUSION}

We introduced {\MethodName}, an omnidirectional stereo matching approach that integrates a large-scale pre-trained monocular depth foundation model into an iterative optimization-based stereo matching framework.
Thanks to a two-stage training strategy, we ensure feature adaptation to omnidirectional stereo matching while preserving the generalization capabilities acquired by the foundation model during its pre-training.
Extensive experiments on the Helvipad dataset demonstrate that {\MethodName} outperforms the previous state of the art by a large margin
across multiple depth and disparity metrics.
Additionally, our model exhibits good generalization capabilities to unseen real-world images and is training sample efficient, highlighting its potential for real-world robotics applications.

\section*{\small{ACKNOWLEDGMENTS}}
\scriptsize{This project was partially supported by the European Research Council (ERC) under the European Union’s Horizon 2020 research and innovation programme (grant agreement No. 866008). Additionally, this work has also been co-funded by the LOEWE initiative (Hesse, Germany) within the emergenCITY center [LOEWE/1/12/519/03/05.001(0016)/72] and was supported by the Deutsche Forschungsgemeinschaft (German Research Foundation, DFG) under Germany’s Excellence Strategy (EXC 3066/1 “The Adaptive Mind”, Project No. 533717223). 
}


\bibliographystyle{IEEEtran}
\bibliography{references}

\end{document}